%% file: main.tex
\documentclass{article}

    \PassOptionsToPackage{numbers, compress}{natbib}
     \usepackage[final]{neurips_2022}

\usepackage[utf8]{inputenc} % allow utf-8 input
\usepackage[T1]{fontenc}    % use 8-bit T1 fonts
\usepackage{hyperref}       % hyperlinks
\usepackage{url}            % simple URL typesetting
\usepackage{booktabs}       % professional-quality tables
\usepackage{amsfonts}       % blackboard math symbols
\usepackage{nicefrac}       % compact symbols for 1/2, etc.
\usepackage{microtype}      % microtypography
\usepackage{xcolor}         % colors

\usepackage{multirow}
\usepackage{graphicx}
\usepackage{amsmath}
\usepackage{amssymb}
\usepackage{amsfonts}
\usepackage{breqn}
\usepackage{caption}
\usepackage{subcaption}
\usepackage{dsfont}
\usepackage{xspace}

\usepackage{wasysym}

\usepackage[accsupp]{axessibility}
\usepackage{bbm}
\usepackage{pifont}
\usepackage{colortbl}
\usepackage{stfloats}
\usepackage{float}
\usepackage{makecell}
\usepackage{wrapfig,lipsum}
\usepackage{tikz}

\usepackage{pythonhighlight}

\usepackage{bbold}

\usepackage[capitalize]{cleveref}
\crefname{section}{Sec.}{Secs.}
\Crefname{section}{Section}{Sections}
\Crefname{table}{Table}{Tables}
\crefname{table}{Tab.}{Tabs.}
\crefname{equation}{Eq.}{Eqs.}

\newcommand*{\myfont}{\bbfamily\selectfont}
\DeclareTextFontCommand{\textmyfont}{\myfont}

\title{\textmyfont{VITA}: Video Instance Segmentation\\via Object Token Association}
\author{%
  Miran Heo\thanks{Both authors contributed equally to this work.} \\
  Yonsei University\\
   \vspace{-6mm}
   \And
   Sukjun Hwang$^*$ \\
   Yonsei University \\
   \vspace{-6mm}
   \AND
   Seoung Wug Oh \\
   Adobe Research \\
   \vspace{-7mm}
   \And
   Joon-Young Lee \\
   Adobe Research \\
   \vspace{-7mm}
   \And
   Seon Joo Kim \\
   Yonsei University \\
   \vspace{-7mm}
  \AND
  \texttt{\{miran, sj.hwang, seonjookim\}@yonsei.ac.kr}
  \And
  \texttt{\{seoh, jolee\}@adobe.com}
}

\begin{document}

\maketitle

\input{sections/0_abstract}
\input{sections/1_introduction}

\input{sections/2_related_works}
\input{sections/3_method}
\input{sections/4_experiments}

\input{sections/5_conclusion}
\clearpage

%\clearpage
\bibliographystyle{plain}
\bibliography{references}

%\input{sections/checklist}

\input{sections/6_supplementary}
\end{document}

%% file: sections/0_abstract.tex
\begin{abstract}
We introduce a novel paradigm for offline Video Instance Segmentation (VIS), based on the hypothesis that explicit object-oriented information can be a strong clue for understanding the context of the entire sequence.
To this end, we propose \textmyfont{VITA}, a simple structure built on top of an off-the-shelf Transformer-based image instance segmentation model.
Specifically, we use an image object detector as a means of distilling object-specific contexts into object tokens.
VITA accomplishes video-level understanding by associating frame-level object tokens without using spatio-temporal backbone features.
By effectively building relationships between objects using the condensed information,
VITA achieves the state-of-the-art on VIS benchmarks with a ResNet-50 backbone: 49.8 AP, 45.7 AP on YouTube-VIS 2019 \& 2021, and 19.6 AP on OVIS.
Moreover, thanks to its object token-based structure that is disjoint from the backbone features, VITA shows several practical advantages that previous offline VIS methods have not explored - handling long and high-resolution videos with a common GPU, and freezing a frame-level detector trained on image domain. Code is available at \url{https://github.com/sukjunhwang/VITA}.

\end{abstract}

%% file: sections/1_introduction.tex
\section{Introduction}
\label{sec:intro}
% 장점들:
% (1) easy and simple
% (2) long sequence (scalability, flexibility, practicality) 
% (3) Fast convergence
% (4) Higher Accuracy
% (5) Eliminate Heuristic Matching for long videos
% (6) Shifted attention
% (7) scene (dense) feature --> object-centric representation feature
The goal of Video Instance Segmentation (VIS) is to predict both mask trajectories and categories of each object belonging to a set of predefined categories.
Numerous studies have attained the goal in a variety of ways, but a notable innovation in terms of accuracy has been achieved by Transformer-based~\cite{Transformer} architectures.
Extending DETR~\cite{DETR} to the video domain, VisTR~\cite{VisTR} made the first attempt to design an end-to-end model that jointly predicts object trajectories with their corresponding segmentation masks.
%By eliminating the post-processing like NMS and \fix{enlarging the scope of the model to the entire video sequence}, VisTR has motivated subsequent studies~\cite{IFC, SeqFormer, Mask2Former-VIS, TeViT} to approach the problem from a new perspective
By adopting this paradigm, subsequent studies~\cite{IFC, SeqFormer, Mask2Former-VIS, TeViT} also tackle the problem in a complete-offline manner: \emph{video-in and video-out}.

The key message from the follow-up approaches~\cite{IFC, SeqFormer, Mask2Former-VIS, TeViT} is to effectively design core interactions between frames.
%For example, VisTR uses self-attention over full spatio-temporal features to exchange information from different frames.
In parallel with recent studies~\cite{SMCA, TSP-RCNN, Deformable-DETR, Mask2Former, Swin} that improve the accuracy in various tasks by localizing the attention scope of Transformer layers, the subsequent VIS methods suggest bounding the attention scope in the encoder~\cite{IFC, TeViT} or the decoder~\cite{SeqFormer}.
% Recent studies~\cite{SMCA, TSP-RCNN, Deformable-DETR, Mask2Former, Swin} have shown that localizing the scope of attention in Transformer layers leads to the better accuracy in various tasks.
%Recently, studies~\cite{SMCA, TSP-RCNN, Deformable-DETR, Mask2Former, Swin} have shown that localizing the scope of attention in Transformer layers leads to the better accuracy in various tasks.
%such dense encoder possibly suffers from over-fitting due to the restricted number of data samples.
%Similarly, follow-up studies in VIS also have suggested that bounding the attention in encoder~\cite{IFC, TeViT} and decoder~\cite{SeqFormer} improves both efficiency and accuracy.
%valuable ideas
%\new{Similarly, succeeding VIS approaches tackle the use of attention in VisTR and suggest to bound the scope of attention in both encoder~\cite{IFC, TeViT} and decoder~\cite{SeqFormer}.}
%Similarly, succeeding VIS approaches suggest that factorizing the attention in encoder~\cite{IFC, TeViT} and decoder~\cite{SeqFormer} improves both efficiency and accuracy.
Specifically, they decompose the global attention by iteratively mixing two phases: intra-frame attention and inter-frame communication.
Interestingly, the temporal interactions between frames are commonly achieved with only a small number of tokens, \textit{e.g.,} memory tokens~\cite{IFC, TeViT}, messenger tokens~\cite{TeViT}, and instance queries~\cite{SeqFormer}.
%Interestingly, the common means for their temporal interactions is a small number of tokens, \textit{e.g.,} memory tokens~\cite{IFC, TeViT}, messenger tokens~\cite{TeViT}, and instance queries~\cite{SeqFormer}.
% RELATED WORK에서 token이 각각 뭔지 간단하게 언급
As a result, the question arises: ``what information is important to understand a video?''
%``which information should be distilled?''
%\emph{which} tokens can effectively build temporal communication, and \emph{where?}''
%As a result, the question arises: \new{\emph{which information should be distilled, where and how?}}
%Although \new{these methods} demonstrate their effectiveness and set compelling performance on benchmarks, the underlying structure hinders their scalability to handle long and high-resolution videos in an end-to-end manner: the backbone features are \fix{strongly involved} in the overall structure as illustrated in \cref{fig:teaser} (b).

In this paper, we introduce \underline{\textbf{V}}ideo \underline{\textbf{I}}nstance Segmentation via Object \underline{\textbf{T}}oken \underline{\textbf{A}}ssociation (\textbf{VITA}), a new offline VIS paradigm which suggests that a video can be effectively understood from a collection of object-centric tokens.
%understands a video using tokens embedded with object-centric information}.
%\new{which understands a video using tokens after having fully embedded with object-centric information}.
Existing offline methods~\cite{VisTR, IFC, SeqFormer, Mask2Former-VIS, TeViT}  (\cref{fig:teaser} (b)) localize objects in multiple frames by iteratively referring to dense spatio-temporal backbone features.
However, such methods show difficulties in handling long sequences as the myriad of dense reference features hinders the Transformer layers from retrieving relevant information.
%However, such methods show difficulties in retrieving relevant information among the myriad of reference tokens where the number further increases .
With the motivation to devise an effective method for the long-range understanding, we obtain clues from the traditional tracking-by-detection paradigm (\cref{fig:teaser} (a)) and make two hypotheses: 1) an image object detector can fully embody the context of an object into a feature vector (or a token); and 2) a video can be represented by the relationship between the objects.

In this regard, VITA aims to parse an input video from the collection of object tokens without the necessity of referencing dense spatio-temporal backbone (\cref{fig:teaser} (c)).
%\new{With the structure disjoint from the detector, VITA achieves notable improvements in the accuracy.}
Given the compactness of the token representation, VITA can collect the object tokens over the whole video and directly analyzes the collection using Transformer layers. 
This unique design enables the complete-offline inference (i.e., video-in and video-out) even for extremely long videos.
This also facilitates building relationships between every detected object and successfully achieves global video understanding.
As a result, VITA achieves state-of-the-art performance on various VIS benchmarks.

We evaluate VITA on three popular VIS benchmarks, YouTube-VIS 2019 \& 2021~\cite{MaskTrackRCNN} and OVIS~\cite{OVIS-Dataset}.
With ResNet-50~\cite{ResNet} backbone, VITA achieves the new state-of-the-arts of 49.8 AP \& 45.7 AP on YouTube-VIS 2019 \& 2021, and 19.6 AP on OVIS.
%With Swin-L~\cite{Swin} backbone, VITA sets a new record of 63.0 on YouTube-VIS 2019.
Above all, VITA outperforms the previous best approaches by 5.1 AP for YouTube-VIS 2021, which contains more complicated and long sequences than YouTube-VIS 2019.
% From the results, we suggest that the direct use of object-oriented information is more effective in dealing with such challenges than using unrefined information from spatio-temporal backbone. 
VITA is the first offline method that presents the results on OVIS benchmark that consists of long videos (the longest video has 292 frames) using a single 12GB GPU.

In addition to the performance, the design of VITA have several practical advantages over the previous offline VIS methods.
It can handle long and high-resolution videos so it does not require heuristics for associating clip-level results. 
VITA can process 1392 frames at once regardless of video resolution using a single 12GB GPU which is 11 times longer than IFC~\cite{IFC}.
Moreover, VITA can be trained on top of a parameter-frozen image object detector without sacrificing the performance much. 
% In this way, one can enjoy video negligible additional parameters
This property is especially useful for the applications that cannot afford to store separated image and video instance segmentation models. 
VITA takes only 6\% additional parameters to extend the Swin-L detector.

% Finally, we demonstrate strong practicality of VITA under highly challenging conditions in which existing methods have not been tested on.
% First, we test the flexibility of modulable architecture by freezing the image detector pretrained on COCO~\cite{COCO}.
% Notably, even if VITA is only fine-tuned to the VIS domain while the detector is frozen, VITA achieves 40.9 AP on the YouTube-VIS 2019 benchmark.
% Second, we test the scalability of the decomposed architecture by feeding long and high-resolution sequences to VITA with a common GPU.
% [ experiment result ].
% VITA shows the potential of a complete offline method being able to process long and high-resolution videos on a light GPU.%, not only on heavy GPU. 
% We hope that our method expands the scope of the offline VIS research to a real-world applications beyond the benchmark.

\input{figure/teaser}

%% file: figure/teaser.tex
\begin{figure}
    \centering
    \includegraphics[width=1.0\linewidth]{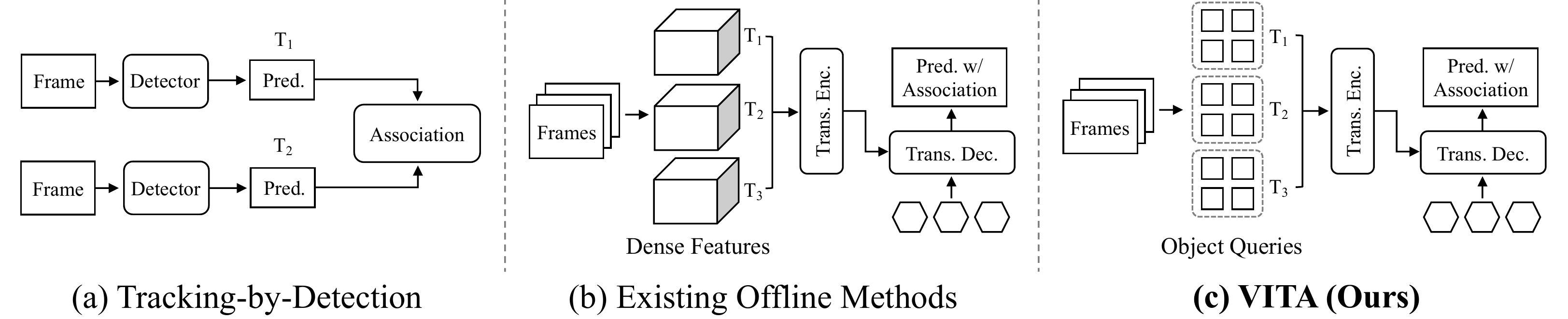}
    %\vspace{-6mm}
    \caption{
    (a) Early-stage VIS methods divide the problem into two components, detection and association. %, following the paradigm of tracking-by-detection.
    (b) To alleviate the context-limited structure, complete-offline methods jointly track and segment instances in an end-to-end manner by employing dense spatio-temporal features.
    (c) On the other hand, our VITA is a new paradigm that directly leverages object queries for offline VIS.
    }
    \label{fig:teaser}
    %\vspace{-5mm}
\end{figure}

%% file: sections/2_related_works.tex
\section{Related Works}
% We divide VIS approaches into two groups in terms of the scope of videos being processed by a model.
% \textit{\fix{Online VIS}} approaches first predict individual tracklets within a local range window consisting of a single or a few frames.
% After obtaining results from adjacent windows, they stitch individual tracklets by a hand-crafted or a learnable matching algorithm.
% On the other hand, \textit{\fix{Offline VIS}} architectures are designed with the motivation of reasoning mask trajectories through a whole video sequence at once. 
% Offline VIS methods have demonstrated the state-of-the-art performance by reducing the error propagation when aggregating tracklets.

\label{sec:related}

\textbf{Online VIS } approaches first predict individual tracklets within a local range window consisting of a single or a few frames.
After obtaining results from adjacent windows, they associate individual tracklets of same identities by a hand-crafted or a learnable matching algorithm.
MaskTrack R-CNN~\cite{MaskTrackRCNN} sets the groundwork for VIS research by proposing a simple tracking branch added on a two-stage image instance segmentation model~\cite{MaskRCNN}.
The methods~\cite{SipMask, CrossVIS, SGNet} that follow the tracking-by-detection paradigm (\cref{fig:teaser} (a)) measure the similarities between per-frame predictions, then employ an association algorithm.
% using the embeddings and locations of each instance.
%Following the tracking-by-detection paradigm (\cref{fig:teaser} (a)), the similarities measured from the embeddings and locations of each instance, are used for the associating predictions during inference.
%\cite{SipMask, CrossVIS, SGNet} also follow the paradigm that learns the object relationship between two frames and associates them frame-by-frame. 

To deploy temporal context from multiple frames, per-clip methods \cite{stem-seg, MaskProp} design an architecture of predicting tracklets within a local window and stitching the tracklets sequentially in a near-online manner.
% 온라인으로 동작하지만 과거의 정보를 더 효율적으로 활용하기 위해 프로퍼게이션 도입.
Propagation-based methods~\cite{CompFeat, PCAN, VISOLO} devise a paradigm that conjugates rich previous information stored in memories to facilitate online applications.
EfficientVIS~\cite{EfficientVIS} introduces correspondence learning between adjacent tracklet features and successfully runs in a cascaded manner which eliminates the hand-crafted tracklet association.
% 몇장을 처리하는지에 대한 차이는 있지만 이어 붙이겠다는 의도는 같음. 에러가 전파될 가능성이 큼.

\textbf{Offline VIS } architectures are proposed with the motivation of predicting mask trajectories through a whole video sequence at once. 
VisTR~\cite{VisTR} successfully extends DETR~\cite{DETR} to the VIS domain, introducing a new paradigm of jointly tracking and segmenting instances. % by modifying its image-multi-frame sequence.
However, its dense self-attention over the spatio-temporal inputs leads to explosive computations and memories.
With the motivation of relaxing the heavy computation of VisTR, IFC~\cite{IFC} adopts memory tokens to the Transformer encoder and decodes clip-level object queries.
% They prove that the efficient communication between context-representative tokens is sufficient rather than the dense self-attention of the spatio-temporal pixel-level feature.
By setting the frame-level encoder to be independent and adopting the decoder of IFC, Mask2Former-VIS~\cite{Mask2Former-VIS} records considerable performance on benchmarks by taking the advantage of its mask-oriented representation~\cite{Mask2Former}. %Specifically, They let the frame-level encoder independent and adopts the decoder of IFC.
% Mask2Former-VIS~\cite{Mask2Former-VIS} extends a strong transformer-based universal image segmentation model~\cite{Mask2Former} to the VIS task.
% Letting frame-level encoder independent and adopting the decoder of IFC, Mask2Former-VIS records considerable performance on benchmarks taking advantage of its segmentation-oriented representation.
% Utilizing iterative refinement of bounding boxes from Deformable DETR~\cite{Deformable-DETR}, SeqFormer~\cite{SeqFormer} further take decompose clip-level queries to frame 
TeViT~\cite{TeViT} proposes a new backbone that efficiently exchanges temporal information internally based on Vision Transformers~\cite{ViT} instead of the frame-wise CNN backbone.
SeqFormer~\cite{SeqFormer} decomposes the decoder to be frame-independent, while building communication between different frames using instance queries that are used for frame-wise detection.
All these studies achieve promising performance by referring to dense backbone features (\cref{fig:teaser} (b)).
%\fix{While these offline studies achieve a promising performance by reducing the error propagation when aggregating tracklets, its dense token-based architecture (\cref{fig:teaser} (b)) forbids  the processing of long and high-resolution videos in a complete-offline fashion.}
On the other hand, our VITA suggests a new offline VIS paradigm that directly interprets a video from the collection of object tokens (\cref{fig:teaser} (c)).

\textbf{Global trackers} that aim to associate frame-level predictions across an entire sequence as a whole are studied in the Multiple Object Tracking (MOT) community.
%\textbf{Global trackers } that explore object association algorithms for stitching frame-level predictions across an entire sequence are studied in the Multiple Object Tracking (MOT) community.
% In the Multiple Object Tracking (MOT) community, some explore object association algorithms for the given frame-level detection across entire sequence.
Conventional approaches formulate the problem as a graph optimization -- interpreting each detection as a node and considering the edges as possible connections between the nodes~\cite{zhang2008global, Lif_T, MPN, LPC}.
Different from existing methods, GTR~\cite{GTR} introduces a Transformer-based architecture that receives queries, then explicitly searches for the predictions with the same identities.
%assigns unique objects to a video clip when high-confidence foreground objects are provided by the two-stage image detector.
Similarly, a recent method~\cite{SetClassifier} proposes a set classifier that classifies the category of each tracklet by globally aggregating information from multiple frames.

%% file: sections/3_method.tex
\section{Method}
\label{sec:methods}
% \new{In this section, we first briefly describe the pipeline of mask-based set prediction detectors~\cite{MaXDeepLab, IFC, Mask2Former}.}
In this section, we first give a brief overview of Mask2Former~\cite{Mask2Former}, a frame-level detector for VITA.
Then, we introduce the architecture of our proposed VITA, which is built on top of Mask2Former.
Finally, we describe how VITA handles extremely long videos in a complete-offline manner. %an end-to-end manner.
%Then we describe the overall architecture and pipeline of our proposed VITA.
\input{figure/main}

\subsection{Frame-level Detector}
\label{sec:framedetector}
In this paper, we adopt Mask2Former~\cite{Mask2Former} for the frame-level detector which directly localizes instances using masks without the necessity of bounding boxes.
Following the set prediction mechanism of DETR~\cite{DETR}, the frame-level detector parse an input image $H \times W$ using $N_f$ object queries, which we call \textit{frame queries} ($f \in \mathbb{R}^{C \times N_f}$) throughout this paper.
% As shown in~\cref{fig:main}, DETR-based~\cite{DETR} detectors~\cite{IFC, Mask2Former, MaXDeepLab} parse an input image $H \times W$ using $N_f$ object queries, which we call \textit{frame queries} ($f$) throughout this paper.
%use $N_f$ object queries to parse an input image $H \times W$.
Having the spatially encoded features to be decoded by the frame queries through a Transformer decoder, each object in the image gets represented as a $C$-dimensional vector.
Then, the frame queries are used for both classifying and segmenting their matched objects where the predictions are also used for auxiliary supervision for VITA.
% \textcolor{red}{Then, the frame queries are used for both classifying and segmenting their matched objects.}
% \new{In this paper, we adopt Mask2Former~\cite{Mask2Former} for the frame-level detector which directly localizes instances without the necessity of bounding boxes.}
Specifically, the frame-level detector generates two features for the frame-level predictions: 1) dynamic $1\times1$ convolutional weight from the frame queries; 2) per-pixel embeddings $\mathcal{M} \in \mathbb{R}^{C \times \frac{H}{S} \times \frac{W}{S}}$ from the pixel decoder, where $S$ is the stride of the feature map.
Finally, the detector segments objects by applying a simple dot product between the two embeddings.
% Specifically, the frame-level detector generates two additional features: 1) mask embeddings $w_f \in \mathbb{R}^{C \times N_f}$ from the frame queries; 2) per-pixel embeddings $\mathcal{M} \in \mathbb{R}^{C \times \frac{H}{S} \times \frac{W}{S}}$ from the pixel decoder, where $S$ is the stride of the feature map.
% Finally, the detector segments objects by applying a simple dot product between the two embeddings.

% Given an input image $H \times W$, recent mask-based set prediction detectors~\cite{IFC, Mask2Former, MaXDeepLab} directly group and classify segments.
% Having the input be encoded using backbones and transformer encoders~\cite{Transformer, Deformable-DETR}, they object-centrically decode the image using object queries $\widehat{f} \in \mathbb{R}^{C \times N_f}$, where $C$ and $N_f$ are the number of channels and object queries, respectively.
% All embedded object queries $f$ are used predict categories and generate dynamic convolutional weights $w_f \in \mathbb{R}^{C \times N_f}$.
% The detectors also integrate multi-level features to spatially decode the image into mask features $\mathcal{M} \in \mathbb{R}^{C \times \frac{H}{S} \times \frac{W}{S}}$, where $S$ is the stride of the feature map.
% From a simple matrix multiplication between $w_f$ and $\mathcal{M}$, each object query obtains a mask prediction.
% Over the $N_f$ predictions, Hungarian algorithm~\cite{Hungarian, DETR} is used for both finding the optimum combination to the ground-truth and eliminating heuristics such as NMS.

\subsection{VITA}
\label{sec:vita}
We now propose the novel end-to-end video instance segmentation method VITA, which can be largely divided into three phases (\cref{fig:main}).
First, VITA operates on top of the frame-level detector~\cite{Mask2Former} in a complete frame-independent manner; no inter-computation between frames is involved.
Then, the frame queries that hold object-centric information are collected throughout the whole video and they embed video-level information by building communications between different frames using Object Encoder.
Finally, Object Decoder aggregates information from the frame queries to video queries, which are eventually used for predicting categories and masks of objects in videos at once.
%Finally, video queries aggregate information from the frame queries through Object Decoder in order to predict categories and masks at once.

%\subsection{Model architecture}
\paragraph{Input of VITA.}
% Input이 들어오고 Base Detector (Mask2Former)를 거쳐서 frame-wise output이 나오는 과정 간략히 설명
Given an input video of $T$ frames, the frame-level detector executes frame-by-frame as previously explained.
Among a number of intermediate embeddings that are generated by the detector, the only features that are used by VITA are 1) frame queries $\{f^t\}_{t=1}^{T} \in \mathbb{R}^{C \times T \times N_f}$ which hold object-centric information; and 2) per-pixel embeddings $\{ \mathcal{M}^t \}_{t=1}^{T} \in \mathbb{R}^{C \times T \times \frac{H}{S} \times \frac{W}{S}}$ from the pixel decoder.

\paragraph{Object Encoder.}
\begin{wrapfigure}{r}{0.45\textwidth}
  \begin{center}
    \vspace{-6mm}
    \includegraphics[width=0.43\textwidth]{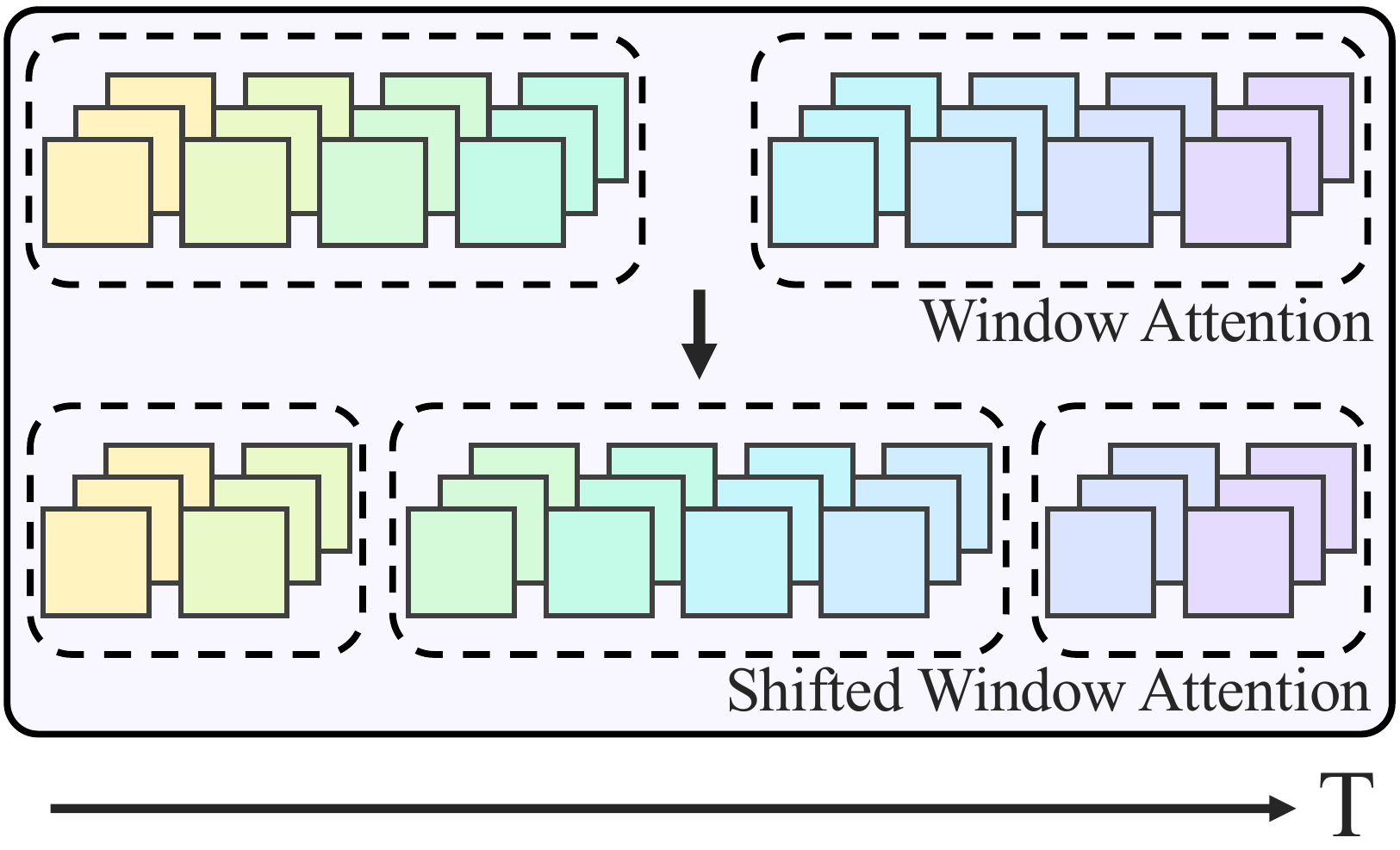}
  \end{center}
  \vspace{-3mm}
  \caption{
    Illustration of an Object Encoder layer.
    Blocks with dashed line are local windows, and $\Square$ indicates an object token.
  }
  \vspace{-5mm}
  \label{fig:clip_encoder}
\end{wrapfigure}
After the frame-wise detector distills the object-wise context into the frame queries, Object Encoder aims to build temporal communication by employing self-attention along the temporal axis.
First, Object Encoder gathers frame queries from all frames and converts them to object tokens through a linear layer.
However, a naive self-attention over the whole $TN_f$ object tokens is not applicable when processing long videos due to the quadratic computational overhead of Transformers.
%However, a naive self-attention is not applicable to the entire \FOQ as a whole due to the inherent characteristic of transformers: the quadratic computational overhead.
Inspired by Swin Transformer~\cite{Swin}, we adopt window-based self-attention layers that shift along the temporal dimension.
As illustrated in~\cref{fig:clip_encoder}, Object Encoder initially partitions object tokens $\{f^t\}_{t=1}^{T}$ to the temporal axis with local windows of size $W$ without an overlap.
By alternatively shifting the windows, object tokens from different frames can exchange object-wise information which allows VITA to both effectively and efficiently handle long sequences.

\paragraph{Object Decoder and Output heads.}
Two limitations of previous offline VIS methods~\cite{VisTR, IFC, Mask2Former-VIS} are the ineffectiveness in handling dynamic scenes and the inability of processing long videos.
For example, such methods obtain high accuracy when dealing with static and short videos (YouTube-VIS 2019~\cite{MaskTrackRCNN}), but struggle to track objects or executes end-to-end on benchmarks with dynamic and long videos (YouTube-VIS 2021~\cite{MaskTrackRCNN} and OVIS~\cite{OVIS-Dataset}).
%For example, such methods show great results in YouTube-VIS 2019~\cite{MaskTrackRCNN} benchmark which is mostly composed of static and short videos.
%However, they fail either accurately tracking objects or executing end-to-end when dynamic and long videos are given (YouTube-VIS 2021~\cite{MaskTrackRCNN} and OVIS~\cite{OVIS-Dataset}).
Both limitations are mainly caused by the decoder, which parses object contexts directly from dense spatio-temporal features.
As recent studies~\cite{SMCA, TSP-RCNN, Deformable-DETR} suggest, typical Transformer decoders show difficulties in retrieving relevant information from global context.
In the video domain, the number of backbone features being referred to proportionally increases with the number of frames.
Therefore, when handling extremely long videos, the countless reference tokens result in both imprecise information retrieval and intractable peak memories.

% All previous DETR-based offline VIS methods~\cite{VisTR, IFC, Mask2Former-VIS} decode object-level information from spatio-temporal backbone features into video object queries.
% Recent studies~\cite{SMCA, TSP-RCNN, Deformable-DETR} have suggested that typical transformer decoder layers find it difficult to retrieve relevant information from global context.
% The inefficiency inevitably becomes more severe in the video domain as the number of tokens being refer to proportionally increases by the number of frames.
% In addition, % long video 안되는 이유

For the solution to the problem, we suggest Object Decoder which extracts information from the object tokens, not the spatio-temporal backbone features.
Implicitly embedding the context of objects, object tokens can provide sufficient instance-specific information without the interference of dense backbone features.
%We hypothesize that \FOQ are feasible to implicitly represent the characteristics of objects and thereby video-level context information can be retrieved from these compact features.
Specifically, we employ $N_v$ trainable video queries $v \in \mathbb{R}^{C \times N_v}$ to decode object-wise information from all object tokens $\{f^t\}_{t=1}^{T}$ that are collected from all $T$ frames.
Receiving much condensed input over naively taking dense spatio-temporal features, Object Decoder effectively captures video contexts and aggregates relevant information into the video queries.
As a result, Object Decoder shows fast convergence speed while achieving high accuracy.
Furthermore, the compact input greatly saves memories, thus facilitates processing long and high-resolution videos.
%Taking $\{f^t\}_{t=1}^{T}$ as keys and values, Object Decoder aggregates relevant information into \COQ to embed video-level context per an object over multiple frames.

% The simple strategy of dramatically reducing the number of reference tokens leads to two great advantages.
% First, processing long and high-resolution videos in an end-to-end manner becomes possible by saving huge amount of memories.
% Second, the smaller number of reference tokens makes it easier for the model to capture the object context over videos, thus accelerates the network convergence while improving the accuracy.

% Clip Decoder links the output of Clip Encoder -- spatio-temporally embedded \FOQ -- by the identities of objects.
% Clip Decoder aggregates information that are embedded in the frame-level object queries
% Through the Clip Encoder layers, the potential objects of each frame exchanges their embeddings temporally.
% To identify same instance across frames, we employ learnable video object queries $v \in \mathbb{R}^{C \times N_v}$, where $C$ and $N_v$ are the number of channels and object queries, respectively.

% Unlike most previous methods that takes spatio-temporal backbone features as keys and values to decode video object queries~\cite{VisTR,IFC,Mask2Former-VIS}, VITA uses output frame-level object features from Clip Encoder.
% In other words, VITA can discard all heavy intermediate features but the frame-level object features $\{f^t\}_{t=1}^{T}$ and mask features $\{\mathcal{M}^t\}_{t=1}^{T}$.

From the decoded video queries $v$, VITA returns final predictions $z=\{(p_i, m_i)\}^{N_v}_{i=1}$ using two output heads similar to IFC~\cite{IFC}; the class head and the mask head.
The class head is a single linear classifier, which directly predicts class probabilities $p \in \mathbb{R}^{N_v \times (K+1)}$ of each video query, where $K+1$ is the number of categories including an auxiliary label ``no object'' ($\varnothing$).
The mask head dynamically generates mask embeddings $w_v \in \mathbb{R}^{C \times N_v}$ per a video query, which corresponds to the tracklet of an instance over all frames.
Finally, the predicted mask logits $m \in \mathbb{R}^{N_v \times T \times H \times W}$ can be obtained from a matrix multiplication between $w_v$ and $\{\mathcal{M}^t\}_{t=1}^{T}$.
%Finally, VITA returns video-level outputs $z=\{(c_i, m_i)\}^{N_v}_{i=1}$.
%dynamically generated convolutional weights from video object queries are applied on frame-level mask features $\mathcal{M}$

\subsection{Clip-wise losses}
\paragraph{Instance matching.}
\begin{wrapfigure}{r}{0.53\textwidth}
  \begin{center}
    \vspace{-7mm}
    \includegraphics[width=0.53\textwidth]{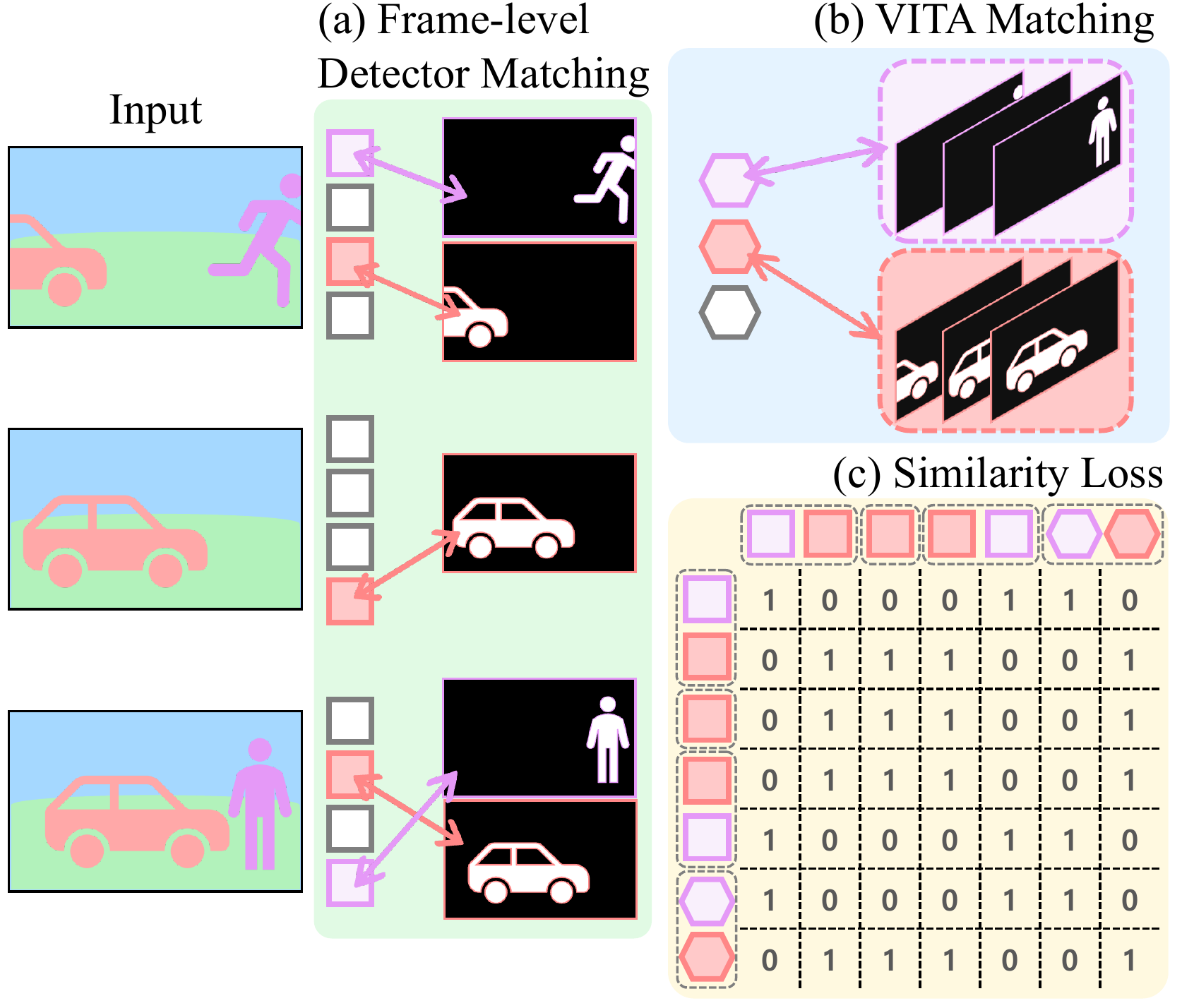}
  \end{center}
  \vspace{-4mm}
  \caption{
    Similarity loss. $\hexagon$ and $\Square$ indicate video query and frame query, respectively.
    Same color represents same GT instance ID.
  }
  \vspace{-8mm}
  \label{fig:sim_loss}
\end{wrapfigure}
We search for optimal pair indices between the predictions from VITA and $G_v$ ground-truth to remove post-processing heuristics such as NMS.
First, we calculate costs from all possible pairs using the cost function of Mask2Former~\cite{Mask2Former} with a simple extension of mask-related costs to the temporal axis~\cite{IFC}.
%With a simple extension of mask-related costs to the temporal axis~\cite{IFC}, we adopt the cost function of~\cite{Mask2Former}.
%Following the cost function of~\cite{Mask2Former}, we compute matching cost between $N_v$ outputs from VITA and $K_v$ ground-truth objects.
Then, from $N_{v} \times G_v$ costs of pairs, we follow DETR~\cite{DETR} and use Hungarian algorithm~\cite{Hungarian} for the optimal matching as shown in~\cref{fig:sim_loss} (b).

%We adopt the cost function of~\cite{Mask2Former} with simple extension of mask-related costs to the temporal axis~\cite{IFC}.
%Then, we use Hungarian algorithm~\cite{Hungarian} to find optimal paired indices $\sigma$ from the cost matrix.

\paragraph{Similarity loss.}
Inspired by the initial VIS approach (MaskTrack R-CNN~\cite{MaskTrackRCNN}) where the similarity loss is adopted to track instances at different frames, we train video queries and frame queries to be clustered in the latent space by their identities.
As shown in~\cref{fig:sim_loss} (a), our adopted frame-level detector~\cite{Mask2Former} also searches for paired indices between $N_{f}$ frame-wise predictions and $G_{f}^{t}$ ground-truth objects at each $t^{\text{th}}$ frame.
%As illustrated in~\cref{sec:framedetector}, VITA runs on top of the frame-level detector~\cite{Mask2Former}, which also searches for paired indices between $N_{f}$ frame-wise predictions and $G_{f}^{t}$ ground-truth objects at each $t^{\text{th}}$ frame.
%Inspired by the initial VIS approach (MaskTrack R-CNN~\cite{MaskTrackRCNN}) where similarity loss is adopted to track instances at different frames, we train video queries and frame queries to be clustered by their identities.
%As shown in~\cref{fig:sim_loss} (a) and (b), a ground-truth gets optimally matched to a query using Hungarian algorithm.
The frame queries and the video queries that are matched to ground-truths get collected and we embed the collection through a linear layer.
Then, we measure the similarity of all possible pairs using a simple matrix multiplication.
Finally, as shown in~\cref{fig:sim_loss} (c), binary cross entropy is used to compute $\mathcal{L}_{sim}$ between the predicted similarities and the ground-truth where annotated to $1$ for pairs of equal identities and $0$ for vice-versa.
%\fix{0516 paragraph review}
%where pairs of an equal identity is annotated 
%Having collected the queries that are matched to ground-truths, we embed the collection through a linear layer and compute similarity of all pairs using a simple matrix multiplication.

\paragraph{Total loss.}
We attach the proposed module VITA on top of the frame-level detector, and all components of the model get trained end-to-end.
Note that not only video-level outputs from VITA are used for the loss computation, but also per-frame outputs from the frame-level detector get involved.
Specifically, we use $\mathcal{L}_{f}$ from~\cite{Mask2Former} to calculate loss from the per-frame outputs to frame-wise ground-truth.
%$\mathcal{L}_v$ is calculated from the outputs of VITA $z$ using the loss function which extends that of~\cite{Mask2Former} to the temporal axis as similar to~\cite{IFC}
%$\mathcal{L}_v$ is the extension of $\mathcal{L}_{f}$ to the temporal axis as similar to~\cite{IFC} which is calculated from the outputs of VITA $z$.
Extending the loss function of~\cite{Mask2Former} to the temporal axis as similar to~\cite{IFC}, we use outputs from VITA $z$ to calculate the video-level loss $\mathcal{L}_v$.
Finally, we integrate all losses together as follows: $\mathcal{L}_{total} = \lambda_{v}\mathcal{L}_{v} + \lambda_{f}\mathcal{L}_{f} + \lambda_{sim}\mathcal{L}_{sim}$.

% Using the outputs from VITA $z$, we calculate $\mathcal{L}_{v}$ with a loss function 
% During training, we calculate $\mathcal{L}_{v}$ from the outputs from VITA $z$
% not only the outputs from VITA are used to calculate $\mathcal{L}_{v}$, but also the per-frame outputs from the frame-wise detector 
% integrate losses from both VITA ($\mathcal{L}_{v}$) and the frame-wise detector~\cite{Mask2Former} ($\mathcal{L}_{f}$) to train end-to-end.
% The loss function of VITA is equal to the cost function which extends $\mathcal{L}_{f}$ to the temporal axis~\cite{IFC}.
% Finally, the final loss is a combination of 
% we use the same cost function for calculating the loss $\mathcal{L}_{vd}$.
% In addition, we also integrate the losses of the frame-wise detector~\cite{Mask2Former} $\mathcal{L}_{det}$ and train end-to-end.
%In addition, we use $\mathcal{L}_{mask}$ of~\cite{Mask2Former} with the temporal extension

%% file: figure/main.tex
\begin{figure}
  \centering
  \includegraphics[width=1.0\linewidth]{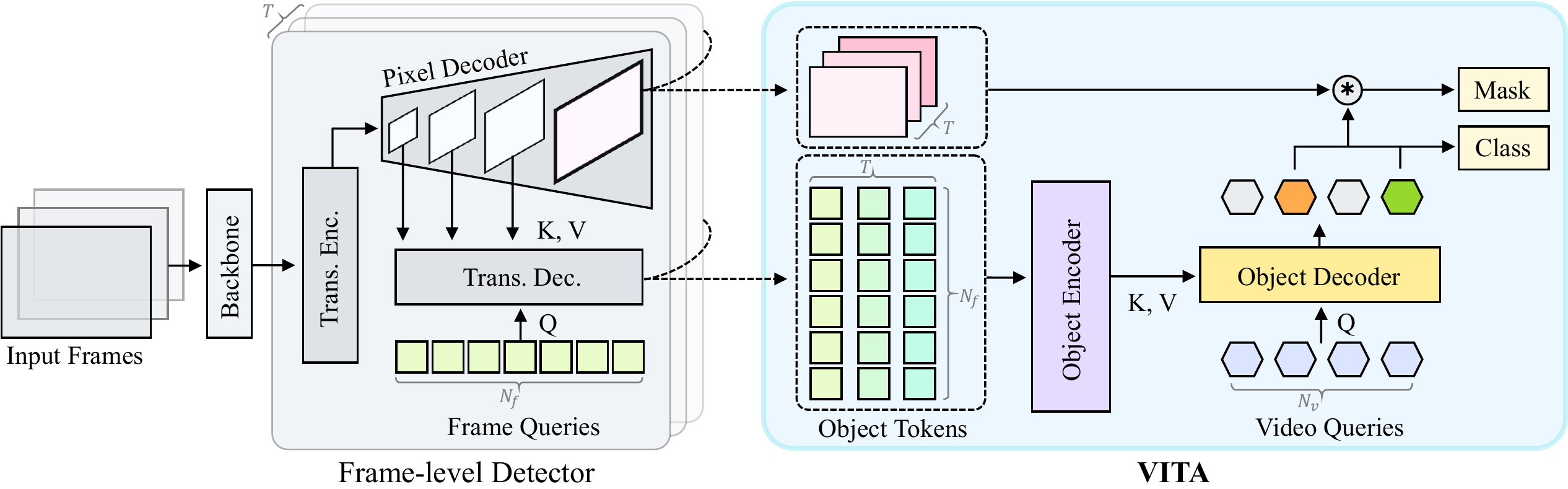}
  %\vspace{-5mm}
  \caption{
  %TODO: CNN->backbone; N_f and N_v; hat(f) and f; Mask feature M;
  VITA takes only mask features and frame queries that are independently decoded by the frame-level detector for entire video sequence.
  By directly constructing temporal interactions between frame queries that encapsulate rich object-aware knowledge in spatial scenes, VITA yields mask trajectories with corresponding categories in an end-to-end manner.
  }
  \label{fig:main}
  %\vspace{-5mm}
\end{figure}

%% file: sections/4_experiments.tex
\section{Experiments}
\label{sec:experiments}
\subsection{Datasets}
\label{sec:datasets}
%We evaluate our method on three VIS benchmarks: YouTubeVIS-2019/2021~\cite{MaskTrackRCNN} and OVIS~\cite{OVIS-Dataset}.
% Before discussing the experimental results, we briefly review the major characteristics of each dataset.

\paragraph{YouTube-VIS 2019.} 
YouTube-VIS 2019~\cite{MaskTrackRCNN} is the first dataset proposed for VIS and contains 40 semantic categories.
Mostly originated from Video Object Segmentation (VOS) datasets, the VIS benchmark has a small number of unique instances (average 1.7 per video for the \texttt{train} set) and the categories of instances appearing in the same video are different in general.
Also, the average length of videos in the \texttt{valid} set is short (27.4 frames), which enables existing complete-offline approaches to load a whole video and infer the benchmark at once.
%thus existing complete offline approaches are able to infer this dataset at once in an end-to-end fashion.

\paragraph{YouTube-VIS 2021.}
In order to address more difficult scenarios, additional videos are included in YouTube-VIS2021 (794 videos for training and 129 videos for validation).
In particular, a greater number of objects with confusing trajectories has been added (average 3.4 per video for the additional videos in the \texttt{train} set).
However, the average length of the additional validation videos is still 39.7 frames, which is not significantly increased compared to YouTube-VIS 2019.

\paragraph{OVIS.}
Under the same definition as YouTube-VIS, OVIS~\cite{OVIS-Dataset} specifically aims to tackle objects with heavy occlusions that are belonging to 25 semantic categories.
In addition to the heavily occluded situation, OVIS has three challenging characteristics that are distinct from the YouTube-VIS datasets.
First, although it has fewer categories than YouTube-VIS, much more instances appear in a single video (average 5.9 per video for the \texttt{train} set).
Second, the instances with the same categories in the same video have almost similar appearances, thus approaches that rely heavily on visual cues often struggle to predict accurate trajectories.
Finally, the average length of videos for the \texttt{valid} set is 62.7 frames (the longest video has 292 frames) which is much longer than that of YouTube-VIS.
Therefore, not only do previous approaches show relatively low accuracy, but all existing complete-offline VIS methods are not feasible to infer OVIS without hand-crafted association algorithms.

%------------------------------
\input{tab/ytvis2019}
%------------------------------

\subsection{Implementation Details}
\label{sec:impl_details}
Our method is implemented on top of \texttt{detectron2}~\cite{Detectron2}.
All hyper-parameters regarding the frame-level detector are equal to the defaults of Mask2Former~\cite{Mask2Former}.
The total loss $\mathcal{L}_{total}$ is balanced with $\lambda_v$, $\lambda_f$, and $\lambda_{sim}$ where 1.0, 1.0, and 0.5, respectively.
By default, Object Encoder is composed of three layers with the window size $W=6$, and Object Decoder employs six layers with $N_v=100$ video queries.
Having VITA built on top of Mask2Former, we first train our model on the COCO~\cite{COCO} dataset following Mask2Former.
Then, we train our method on the VIS datasets~\cite{MaskTrackRCNN, OVIS-Dataset} simultaneously with pseudo videos generated from images~\cite{COCO} following the details of SeqFormer~\cite{SeqFormer}.
%Specifically, a video sample during training is composed of $T=5$ frames, 
During inference, each frame is resized to a shorter edge size of 360 and 448 pixels when using ResNet~\cite{ResNet} and Swin~\cite{Swin} backbones, respectively.
Note that all reported scores in main results and ablation studies are the mean of five runs, and we use the standard ResNet-50~\cite{ResNet} for the backbone unless specified.

%\fix{attention size (training/test), number of queries, additional training dataset, data augmentation same as mask2former, lambda (v, f, sim)}

\subsection{Main Results}
Using the popular VIS benchmarks -- YouTube-VIS 2019 \& 2021~\cite{MaskTrackRCNN} and OVIS~\cite{OVIS-Dataset} -- we compare VITA with state-of-the-art approaches following the standard evaluation metric~\cite{MaskTrackRCNN}.

\paragraph{YouTube-VIS 2019.}
\cref{tab:ytvis2019} shows the comparison on YouTube-VIS 2019 dataset with backbones of both CNN-based (ResNet-50 and 101~\cite{ResNet}) and Transformer-based (Swin-L~\cite{Swin}). % change when SWIN results are added
%We categorize the methods into (near) online and offline.
%We provide experimental results according to the use of sequential results when inference: (near) online and offline.
%If a method provides multiple scores for various window size for test, we report its performance of fully offline execution~\cite{IFC, TeViT}.
Offline methods can take two advantages over (near) online approaches: 1) they have a greater receptive field to the temporal axis, and 2) they can avoid error propagation derived from hand-crafted association algorithms.
As a result, the tendency of offline methods with higher accuracy is clearly shown in the table.
%Taking the advantage of greater receptive field to the temporal axis, it is clearly shown in the table that offline methods tend to have better accuracy than online approaches.
%the overall performance of offline methods are higher than that of (near) online approaches.
%Obviously, the overall performance of offline methods higher than (near) online methods by enlarging temporal receptive field and eliminating error propagation of heuristic matching.
Among the competitive offline models, our VITA sets a new state-of-the-art of 49.8 AP and 51.7 AP using CNN backbones, ResNet-50 and ResNet-101 respectively.
In addition, with Swin-L backbone, VITA achieves 63.0 AP outperforming all existing VIS methods.
%With both ResNet-50 and ResNet-101 backbones, inter alia, VITA sets new state-of-the-art of 49.8 AP and 51.7 AP respectively.

%------------------------------
\input{tab/ytvis2021_ovis}
%------------------------------

\paragraph{YouTube-VIS 2021.}
We compare VITA with state-of-the-art methods on YouTube-VIS 2021 benchmark in~\cref{tab:ytvis2021}. 
%All methods listed in the table but TeViT~\cite{TeViT} use ResNet-50 as a backbone.
%Also,
Note that the longest video in the \texttt{valid} set has 84 frames, thus previous offline methods~\cite{IFC, SeqFormer, Mask2Former-VIS} can infer videos at once with GPUs with large memories.
%there is still no post-processing for previous offline methods~\cite{IFC, SeqFormer, Mask2Former-VIS} because the longest video in validation set has 84 frames which can be handled by using GPUs with large memories.
Above all, VITA achieves the highest accuracy, 45.7 AP, which outperforms the previous state-of-the-art approach~\cite{Mask2Former-VIS} with a huge margin of 5.1 AP.
%surpasses all existing methods. 
%Especially, VITA outperforms the previous state-of-the-art approach with the huge margin: 5.1 AP.
Considering the accuracy gap on YouTube-VIS 2019, the results demonstrate that VITA can effectively handle tricky scenarios, \textit{e.g.,} numerous unique instances with confusing trajectories.
We hypothesize that the object-oriented design of VITA is more effective than typical dense Transformer decoders in addressing such challenging scenes.
%the explicit object-oriented information is more effective at addressing such challenges than extensive information about the scene.

\paragraph{OVIS.}
In~\cref{tab:ytvis2021}, we demonstrate the competitiveness of VITA on the challenging OVIS benchmark.
Due to the considerable lengths of videos -- the longest video has 292 frames -- existing offline approaches~\cite{VisTR, IFC, TeViT, SeqFormer, Mask2Former-VIS} cannot process OVIS benchmark in their original design: video-in and video-out.
%Unlike the results on YouTube-VIS, existing complete-offline approaches cannot process OVIS dataset in an end-to-end fashion because the longest video in validation set is consisted of 292 frames.
To the best of our knowledge, VITA is the first complete-offline approach to evaluate on OVIS \texttt{valid} set.
Thanks to its object token-based structure which is disjoint from backbone features, VITA can process the benchmark \emph{without} any hand-crafted association algorithm.
Moreover, VITA sets a new state-of-the-art performance of 19.6 AP, demonstrating the potential of the complete-offline pipeline in long and complicated scenes.
% of a mechanism to globally associate the fairly long sequences in complicated scenes.

\subsection{Ablation Studies}
We provide a series of ablation studies using a ResNet-50~\cite{ResNet} backbone.
All experiments are conducted on YouTube-VIS 2019~\cite{MaskTrackRCNN} \texttt{valid} set except for \cref{tab:clip_matching} with OVIS~\cite{OVIS-Dataset} \texttt{valid} set.
%and most experiments are conducted on YouTube-VIS 2019 dataset using ResNet-50 as a backbone unless specified.

%------------------------------
\input{tab/window}
%------------------------------

\paragraph{Attention window size.}
\cref{tab:window_size} shows the performance of VITA with varying sizes of shifted attention window $W$ in Object Encoder during inference.
The larger the window, the greater the receptive field for the temporal axis in Object Encoder.
The results suggest that larger window sizes utilize information from multiple frames, which helps Object Encoder understand the context of objects in videos.
We set $W=6$ considering a trade-off between performance and inference scalability.

\paragraph{Maximum number of frames.}
In \cref{tab:max_frame}, we calculate the maximum number of frames that VITA can handle with respect to the various window sizes $W$, and compare it with existing complete-offline VIS methods.
To take into account the general environment, all results are computed using a single 12GB Titan XP GPU.
As shown in results, existing methods have limitations in processing long videos in a \emph{video-in and video-out} manner.
Clearly, the bottleneck of VisTR~\cite{VisTR} is the encoder, where the full spatio-temporal self-attention leads to a tremendous memory usage.
IFC~\cite{IFC} alleviates the computation of VisTR~\cite{VisTR}, achieving a higher number of input frames.
However, IFC makes use of a typical Transformer decoder that visits all dense spatio-temporal features.
Therefore, IFC cannot infer the OVIS~\cite{OVIS-Dataset} benchmark at once which contains a video of 292 frames.
The problem gets aggravated in Mask2Former-VIS~\cite{Mask2Former-VIS} as the scope of the decoder is extended to multiple feature levels~\cite{Mask2Former}.
%not only the decoder visits all dense spatio-temporal features, but also multiple levels.
%On the other hand, the typical Transformer decoder of Mask2Former-VIS~\cite{Mask2Former-VIS} is what decreases the affordable frame number as it visits all dense spatio-temporal backbone features of multiple levels.
%On the other hand, the decoder where all dense multi-level spatio-temporal tokens get decoded at once is the major bottleneck in Mask2Former-VIS~\cite{Mask2Former-VIS}.
%IFC~\cite{IFC} alleviates the computation in VisTR~\cite{VisTR}, achieving higher acceptable frame number.
%However, IFC still suffers from the same bottleneck of Mask2Former~VIS~\cite{Mask2Former-VIS} due to the use of the typical Transformer decoder.
% due to its underlying dense spatio-temporal Transformer decoder.
% Even though IFC~\cite{IFC} directly aims to reduce the heavy computation of encoder architecture, forbids ---.
%Moreover, this becomes much more insufficient as the input frame resolution increases.
On the other hand, VITA presents considerable frame numbers that can be inferred completely offline.
Furthermore, VITA is independent from input frame resolutions as each frame gets summarized into compact object tokens.
%On the other hand, VITA is independent of its input resolution because it only takes object tokens and mask features which are decoded independently from frame-level detector.
With input resolution of $360 \times 640$ and $W=6$, the maximum length of sequence that VITA is able to process in complete-offline is about \emph{11$\times$ longer} than IFC~\cite{IFC}.

\paragraph{Heuristic clip association.}
\cref{tab:clip_matching} shows the results on OVIS \texttt{valid} set of splitting a video into shorter clips and associating clip-wise predictions through heuristic matching.
The length of the clip is set to be less than the average length of videos of OVIS \texttt{valid} set (62.7).
Then, we associate outputs from different clips using mask IoU score as the matching cost. 
We test with two matching algorithms: Greedy and Hungarian. 
%Then we make one frame of overlap between clips, and compute mask IoU of all instance pairs as a matching cost for bipartite matching.
%For the algorithms for bipartite matching, we use both Greedy and Hungarian.
As shown in~\cref{tab:clip_matching}, VITA demonstrates the best performance on the complete-offline inference that use all the video frames at once.

\paragraph{Pruning Tokens.}
In~\cref{tab:pruning}, we investigate the effects of removing redundant frame queries.
From a collection of frame queries, VITA understands the overall context of the given clip.
As only a small portion of the collection is matched to foreground objects, the number of total input frame queries can be reduced.
First, for each frame, we sort frame queries in ascending order by the ``no object'' ($\varnothing$) probability.
Then, we keep only top $rN_{f}$ queries from the sorted list where $r$ is the ratio, and discard the rest.
The accuracy with respect to the ratio $r$ is as shown in~\cref{tab:pruning}.

By setting the ratio $r=0.75$, the accuracy of VITA shows only a marginal degradation in the accuracy ($-0.1$ AP).
This signifies that VITA focuses more on the foreground contexts that are embedded in the frame queries.
Meanwhile, as the quadratic computation in Clip Encoder can be alleviated, VITA can process a much greater number of frames; using the ratio $r=0.75$, the maximum frame number increases from 1392 (\cref{tab:max_frame}) to 2635.

\paragraph{Convergence speed and Similarity loss.}
\cref{fig:convergence} validates our claim of the faster convergence speed and the effectiveness of the proposed Similarity loss.
For a fair comparison, we report average scores and standard deviations of five runs, each trained without pseudo videos, same as Mask2Former-VIS~\cite{Mask2Former-VIS}.
Thanks to its object-centric design, VITA shows faster convergence than Mask2Former-VIS.
Furthermore, the use of Similarity loss leads to an additional accuracy gain of 1.8 AP.
The results demonstrate that the loss mitigates the discrepancies between the embeddings of equal identities, leading to better performance.
% YTVIS
% We follow IFC but faster convergence --> shorter training schedule

\paragraph{Frozen frame-level detector.}
In~\cref{tab:freeze_coco}, we demonstrate the performance of VITA where the frame-level detector is completely frozen.
%1) lack of powerful GPUs; 2) having limited time to train models; and 3) when necessary to keep image segmentation performance of a model while extending it to video domains.
Specifically, while VITA gets trained on YouTube-VIS 2019, the frame-level detector~\cite{Mask2Former} does not get updated from pretrained weights on COCO~\cite{COCO}.
%the frame-level detector~\cite{Mask2Former} does not get updated from pretrained weights on COCO~\cite{COCO} during the training of VITA to YouTube-VIS 2019.
%To explore the flexibility of VITA under practical purposes, we train only VITA by freezing the frame-level detector which is pretrained on COCO dataset.
%\cref{tab:freeze_coco} shows the results with both CNN and Transformer based backbones.
Note that among 40 categories in YouTube-VIS 2019 dataset, only 20 categories overlap with the categories of COCO.
Interestingly, though the frame-level detector remains completely frozen, VITA achieves compelling results with various backbones.
As shown in~\cref{tab:ytvis2019} and~\cref{tab:freeze_coco}, VITA presents a huge practicality as it surpasses all online approaches on top of the ResNet-50 backbone.
This strategy can be beneficial in various scenarios: 1) when the accuracy of image instance segmentation should be kept while extending the network to the video domain, and 2) when having limited time and GPUs to train models.
The strategy can be especially useful in mobile applications that have scarce storage for keeping two separate network parameters for image instance segmentation and video instance segmentation.
With additional 6\% parameters, VITA successfully extends the frozen Swin-L based frame-level detector to the video domain and it achieves great accuracy.

We also provide a brief discussion of our understanding for the large gap in AP. Compared to COCO, we observe that YouTube-VIS dataset is annotated with only a few salient objects as foregrounds. Having weights of the frame-level detector frozen to COCO, the detector cannot adapt to the YouTube-VIS domain and it embeds and interprets more objects in scenes as foregrounds. Therefore, VITA outputs more predictions as a foreground category even if such predictions are not labeled as ground-truths in YouTube-VIS. As a result, it leads to a lower average precision as it comes out with more false positive predictions. On the contrary, the more false positive predictions only slightly affect AR.
% This strategy can be beneficial in various scenarios: 1) when necessary to keep image segmentation performance of a model while extending it to video domains; 2) lack of storage for keeping network parameters; and 3) having limited time and GPUs to train models.
% For the first scenario, the 
% 6\%
%  for segmenting both images and videos

% even if we freeze the classification head of frame-level detector, VITA achieves plausible performance with all backbones.
% Considering that YouTube-VIS 2019 dataset includes 20 categories that overlap with COCO out of 40 categories, VITA ----.
% \fix{here: we need to suggest and discuss various options.}
% This suggests that VITA widens choices. %meaningful.
% For instance, by freezing the detector, the number of parameters used for training can be drastically reduced, so learning can be performed even with a GPU with less memory.
% 빠르게 이미지 detector video로 deploy 하고 싶을 때
% 트레이닝 리소스가 적지만, 성능 높이고 싶을 때

\input{tab/ablations}

\paragraph{Qualitative Results.}
We provide some visualizations of the predictions from VITA and frame-level detector in~\cref{fig:qualitative}.
The qualitative results show that VITA leads to better video instance segmentation qualities compared to the frame-level detector.
Specifically, the frame-level detector mistakenly interprets in to recognize either category or mask of instances that have been largely occluded, while our method successfully recovers it by leveraging the temporal information.

\input{figure/qualitative.tex}

%% file: tab/ytvis2019.tex
\begin{table}
\centering
\caption{
Comparisons on YouTube-VIS 2019.
} % \caption
%\resizebox{\linewidth}{!}
{ %< auto-adjusts font size to fill line
\begin{tabular}{c|l|l|ccccc}
    \toprule
\multicolumn{2}{l|}{Method}                     & Backbone    & AP        & AP$_{50}$ & AP$_{75}$ & AR$_1$    & AR$_{10}$ \\
    \midrule
    \midrule
    \multirow{9}{*}{\rotatebox{90}{(Near) Online}}
    & MaskTrack R-CNN~\cite{MaskTrackRCNN}      & ResNet-50                 & 30.3      & 51.1      & 32.6      & 31.0      & 35.5      \\
    & MaskTrack R-CNN~\cite{MaskTrackRCNN}      & ResNet-101                & 31.8      & 53.0      & 33.6      & 33.2      & 37.6      \\
    %& SipMask~\cite{SipMask}                    & ResNet-50                 & 33.7      & 54.1      & 35.8      & 35.4      & 40.1      \\
    %& STEm-Seg~\cite{stem-seg}                  & ResNet-101                & 34.6      & 55.8      & 37.9      & 34.4      & 41.6      \\
    %& MaskProp~\cite{MaskProp}                  & ResNet-50                 & 40.0      & -         & 42.9      & -         & -         \\
    %& MaskProp~\cite{MaskProp}                  & ResNet-101                & 42.5      & -         & 45.6      & -         & -         \\
    %& SG-Net~\cite{SGNet}                       & ResNet-50                 & 34.8      & 56.1      & 36.8      & 35.8      & 40.8      \\
    %& SG-Net~\cite{SGNet}                       & ResNet-101                & 36.3      & 57.1      & 39.6      & 35.9      & 43.0      \\
    & CrossVIS~\cite{CrossVIS}                  & ResNet-50                 & 36.3      & 56.8      & 38.9      & 35.6      & 40.7      \\
    & CrossVIS~\cite{CrossVIS}                  & ResNet-101                & 36.6      & 57.3      & 39.7      & 36.0      & 42.0      \\
    & PCAN~\cite{PCAN}                          & ResNet-50                 & 36.1      & 54.9      & 39.4      & 36.3      & 41.6      \\
    & PCAN~\cite{PCAN}                          & ResNet-101                & 37.6      & 57.2      & 41.3      & 37.2      & 43.9      \\
    & EfficientVIS~\cite{EfficientVIS}          & ResNet-50                 & 37.9      & 59.7      & 43.0      & 40.3      & 46.6      \\
    & EfficientVIS~\cite{EfficientVIS}          & ResNet-101                & 39.8      & 61.8      & 44.7      & 42.1      & 49.8      \\
    & VISOLO~\cite{VISOLO}                      & ResNet-50                 & 38.6      & 56.3      & 43.7      & 35.7      & 42.5      \\
    
    \midrule
    \multirow{14}{*}{\rotatebox{90}{Offline}}
    & VisTR~\cite{VisTR}                        & ResNet-50                 & 35.6      & 56.8      & 37.0      & 35.2      & 40.2      \\
    & VisTR~\cite{VisTR}                        & ResNet-101                & 38.6      & 61.3      & 42.3      & 37.6      & 44.2      \\
    & IFC~\cite{IFC}                            & ResNet-50                 & 41.2      & 65.1      & 44.6      & 42.3      & 49.6      \\
    & IFC~\cite{IFC}                            & ResNet-101                & 42.6      & 66.6      & 46.3      & 43.5      & 51.4      \\
    & TeViT~\cite{TeViT}                        & MsgShifT                  & 46.6      & 71.3      & 51.6      & 44.9      & 54.3      \\
    & SeqFormer~\cite{SeqFormer}                & ResNet-50                 & 47.4      & 69.8      & 51.8      & 45.5      & 54.8      \\
    & SeqFormer~\cite{SeqFormer}                & ResNet-101                & 49.0      & 71.1      & 55.7      & 46.8      & 56.9      \\
    & SeqFormer~\cite{SeqFormer}                & Swin-L                    & 59.3      & 82.1      & 66.4      & 51.7      & 64.4      \\
    & Mask2Former-VIS~\cite{Mask2Former-VIS}    & ResNet-50                 & 46.4      & 68.0      & 50.0      & -         & -         \\
    & Mask2Former-VIS~\cite{Mask2Former-VIS}    & ResNet-101                & 49.2      & 72.8      & 54.2      & -         & -         \\
    & Mask2Former-VIS~\cite{Mask2Former-VIS}    & Swin-L                    & 60.4      & 84.4      & 67.0      & -         & -         \\
    \cmidrule{2-8}
    &                                           & ResNet-50                 & 49.8      & 72.6      & 54.5      & 49.4      & 61.0      \\
    & VITA (Ours)                               & ResNet-101                & 51.9      & 75.4      & 57.0      & 49.6      & 59.1      \\
    &                                           & Swin-L                    & 63.0      & 86.9      & 67.9      & 56.3      & 68.1      \\
    \bottomrule
    \end{tabular}
} %< \resizebox
\label{tab:ytvis2019}
\end{table}

%% file: tab/ytvis2021_ovis.tex
\begin{table}
\centering
\caption{
Comparisons with ResNet-50 backbone on YouTube-VIS 2021 and OVIS.
$\dagger$ indicates using MsgShifT backbone.
$\ddagger$ indicates using Swin-L~\cite{Swin} backbone.
} % \caption
\resizebox{\linewidth}{!}
{ %< auto-adjusts font size to fill line
\begin{tabular}{@{}l|ccccc|ccccc@{}}
\toprule
\multirow{2}{*}{Method}                 & \multicolumn{5}{c|}{YouTube-VIS 2021}                 & \multicolumn{5}{c}{OVIS}\\
                                        & AP    & AP$_{50}$ & AP$_{75}$ & AR$_{1}$  & AR$_{10}$ & AP    & AP$_{50}$ & AP$_{75}$ & AR$_{1}$  & AR$_{10}$ \\
\midrule
\midrule
MaskTrack R-CNN~\cite{MaskTrackRCNN}    & 28.6  & 48.9      & 29.6      & 26.5      & 33.8      & 10.8  & 25.3      & 8.5       & 7.9       & 14.9  \\
CMaskTrack R-CNN~\cite{OVIS}            & -     & -         & -         & -         & -         & 15.4  & 33.9      & 13.1      & 9.3       & 20.0  \\
STMask~\cite{STMask}                    & 31.1  & 50.4      & 33.5      & 26.9      & 35.6      & 15.4  & 33.8      & 12.5      & 8.9       & 21.3  \\
%SipMask~\cite{SipMask}                  & 31.7  & 52.5      & 34.0      & 30.8      & 37.8      & 10.2  & 24.7      & 7.8       & 7.9       & 15.8  \\
%TraDeS~\cite{TraDeS}                    & -     & -         & -         & -         & -         & 11.4  & 26.5      & 9.4       & 7.0       & 13.8  \\
%STEm-Seg~\cite{stem-seg}                & -     & -         & -         & -         & -         & 13.8  & 32.1      & 11.9      & 9.1       & 20.0  \\
%QueryInst-VIS~\cite{QueryInst-VIS}      & 14.7  & 34.7      & 11.6      & 9.0       & 21.2      \\
CrossVIS~\cite{CrossVIS}                & 34.2  & 54.4      & 37.9      & 30.4      & 38.2      & 14.9  & 32.7      & 12.1      & 10.3      & 19.8  \\
IFC~\cite{IFC}                          & 35.2  & 55.9      & 37.7      & 32.6      & 42.9      & -     & -         & -         & -         & -     \\
VISOLO~\cite{VISOLO}                    & 36.9  & 54.7      & 40.2      & 30.6      & 40.9      & 15.3  & 31.0      & 13.8      & 11.1      & 21.7  \\
TeViT$^\dagger$~\cite{TeViT}            & 37.9  & 61.2      & 42.1      & 35.1      & 44.6      & 17.4  & 34.9      & 15.0      & 11.2      & 21.8  \\
SeqFormer~\cite{SeqFormer}              & 40.5  & 62.4      & 43.7      & 36.1      & 48.1      & -     & -         & -         & -         & -     \\
Mask2Former-VIS~\cite{Mask2Former-VIS}  & 40.6  & 60.9      & 41.8      & -         & -         & -     & -         & -         & -         & -     \\
\midrule
\textbf{VITA (Ours)}                    & \textbf{45.7} & \textbf{67.4} & \textbf{49.5} & \textbf{40.9} & \textbf{53.6} & \textbf{19.6} & \textbf{41.2} & \textbf{17.4} & \textbf{11.7} & \textbf{26.0}\\
\midrule
\midrule
SeqFormer$^\ddagger$~\cite{SeqFormer}   & 51.8  & 74.6      & 58.2      & 42.8      & 58.1      & -     & -         & -         & -         & -     \\
Mask2Former-VIS$^\ddagger$~\cite{Mask2Former-VIS}  & 52.6  & 76.4      & 57.2      & -         & -         & -     & -         & -         & -         & -     \\
\midrule
\textbf{VITA (Ours)$^\ddagger$}                    & \textbf{57.5} & \textbf{80.6} & \textbf{61.0} & \textbf{47.7} & \textbf{62.6} & \textbf{27.7} & \textbf{51.9} & \textbf{24.9} & \textbf{14.9} & \textbf{33.0}\\
\bottomrule
\end{tabular}
} %< \resizebox
\label{tab:ytvis2021}
\end{table}

%% file: tab/window.tex
\begin{table}
	\begin{minipage}[t]{0.48\linewidth}
		\centering
        \caption{
        Impact of local windows of varying sizes in Object Encoder.
        } % \caption
        %\resizebox{\linewidth}{!}
        { %< auto-adjusts font size to fill line
        \begin{tabular}{@{}c|ccccc@{}}
        \toprule
        $W$ & AP    & AP$_{50}$ & AP$_{75}$ & AR$_{1}$  & AR$_{10}$ \\
        \midrule
        \midrule
        %1   & 49.1  & 72.1      & 53.0      & 48.8      & 60.7      \\
        3   & 49.4  & 72.2      & 54.4      & 48.6      & 60.9      \\
        6   & 49.8  & 72.6      & 54.5      & 49.4      & 61.0      \\
        12  & 50.0  & 73.0      & 54.7      & 49.0      & 60.8      \\
        All & 50.1  & 72.4      & 54.7      & 49.0      & 60.6      \\
        \bottomrule
        \end{tabular}
        } %< \resizebox
        %\vspace{-9mm}
        \label{tab:window_size}
	\end{minipage}
	\hfill
	\begin{minipage}[t]{0.48\linewidth}
        \centering
        \caption{
            Maximum number of frames that can be processed at once using a single Titan XP.
        }
        \resizebox{\linewidth}{!}
        { %< auto-adjusts font size to fill line
        \begin{tabular}{@{}cl|cc@{}}
        \toprule
        \multicolumn{2}{l|}{\multirow{2}{*}{Method}}                & \multicolumn{2}{c}{Max Frames}\\
                                        &                           & $360 \times 640$  & $720 \times 1280$\\
        \midrule
        \midrule
        \multicolumn{2}{l|}{VisTR~\cite{VisTR}}                     & 46                & 12    \\
        \multicolumn{2}{l|}{IFC~\cite{IFC}}                         & 123               & 38    \\
        \multicolumn{2}{l|}{Mask2Former-VIS~\cite{Mask2Former-VIS}} & 81                & 20    \\
        \midrule
        \multirow{3}{*}{\shortstack{VITA\\(Ours)}}  & $W = 3$       & \multicolumn{2}{c}{2677}  \\
                                                    & $W = 6$       & \multicolumn{2}{c}{1392}  \\
                                                    & $W = 12$      & \multicolumn{2}{c}{741}   \\
                                                    %& w = 18       & \multicolumn{2}{c}{489}   \\
        \bottomrule
        \end{tabular}
        } %< \resizebox
        \label{tab:max_frame}
        %\vspace{-9mm}
	\end{minipage}
    \hfill

    \begin{minipage}[t]{0.48\textwidth}
        \vspace{-9mm}
        \centering
        \caption{
            Use of different heuristic association algorithms on OVIS \texttt{valid} set.
        }
        % \resizebox{\linewidth}{!}
        {
        \begin{tabular}{@{}c|c|ccc@{}}
            \toprule
            Length               & Algorithm            & AP    & AP$_{50}$ & AP$_{75}$\\
            \midrule
            \midrule
            \multirow{2}{*}{36}  & Greedy               & 18.8  & 39.4      & 17.1\\
                                 & Hungarian            & 18.4  & 38.9      & 16.3\\
            \midrule
            \multirow{2}{*}{48}  & Greedy               & 18.8  & 39.0      & 17.1\\
                                 & Hungarian            & 19.1  & 39.1      & 17.4\\
            \midrule
            All                  & None                 & 19.6  & 41.2      & 17.4\\
            \bottomrule
        \end{tabular}
        }
        \label{tab:clip_matching}
    \end{minipage}
    \hfill
    \begin{minipage}[t]{0.48\linewidth}
        \centering
        \vspace{2.3mm}
        \caption{Pruning tokens by different ratios $r$.}
        \begin{tabular}{@{}l|ccccc@{}}
        \toprule
        $r$         & AP    & AP$_{50}$ & AP$_{75}$ & AR$_{1}$  & AR$_{10}$ \\
        \midrule
        \midrule
        1.0         & 49.8  & 72.6      & 54.5      & 49.4      & 61.0      \\
        0.75        & 49.7  & 72.5      & 54.4      & 48.7      & 61.0      \\
        0.5         & 48.9  & 72.1      & 52.0      & 48.3      & 60.9      \\
        0.25        & 48.1  & 71.6      & 51.6      & 47.4      & 59.8      \\
        \bottomrule
        \end{tabular}
        \label{tab:pruning}
    \end{minipage}
    \hfill
\end{table}

%% file: tab/ablations.tex
\begin{table}
	\begin{minipage}{0.43\linewidth}
		\centering
		\includegraphics[width=\textwidth]{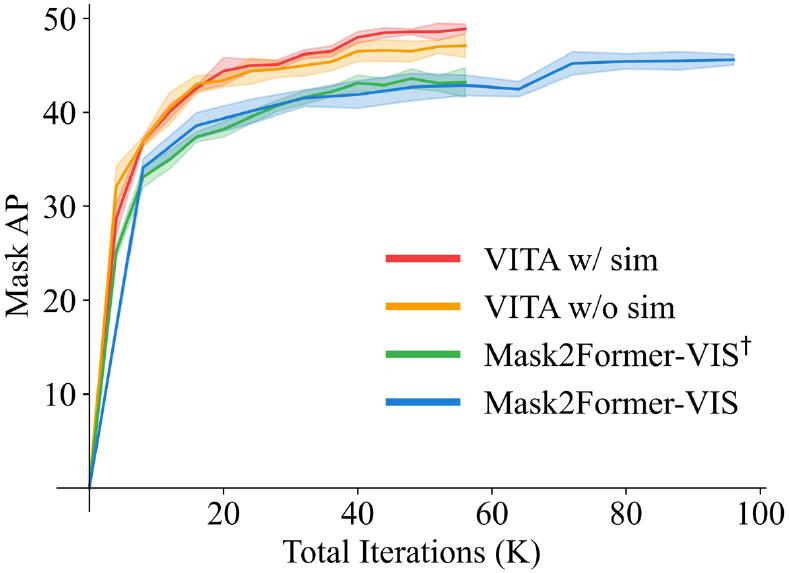}
		\captionof{figure}{
		    Train speed comparison with Mask2Former-VIS~\cite{Mask2Former-VIS}.
            $\dagger$ indicates the same training setup with VITA.
		}
		%\vspace{-5mm}
		\label{fig:convergence}
	\end{minipage}
	\hfill
	\begin{minipage}{0.55\linewidth}
        \centering
        %\vspace{-7mm}
        \caption{
        Results on YouTube-VIS 2019 with freezing detector pretrained on COCO.
        } % \caption
        \resizebox{\linewidth}{!}
        { %< auto-adjusts font size to fill line
        \begin{tabular}{@{}l|c|ccccc@{}}
        \toprule
        Backbone                    & Freeze        & AP    & AP$_{50}$ & AP$_{75}$ & AR$_{1}$  & AR$_{10}$ \\
        \midrule
        \midrule
        \multirow{2}{*}{ResNet-50}  &               & 49.8  & 72.6      & 54.5      & 49.4      & 61.0      \\
                                    & \checkmark    & 40.9  & 61.9      & 44.6      & 43.1      & 53.1      \\
        \midrule
        \multirow{2}{*}{ResNet-101} &               & 51.9  & 75.4      & 57.0      & 49.6      & 59.1      \\
                                    & \checkmark    & 43.2  & 64.4      & 48.7      & 46.1      & 55.9      \\
        \midrule
        \multirow{2}{*}{Swin-L}     &               & 63.0  & 86.9      & 67.9      & 56.3      & 68.1      \\
                                    & \checkmark    & 53.4  & 75.9      & 58.7      & 51.9      & 64.3      \\
        \bottomrule
        \end{tabular}
        } %< \resizebox
        \label{tab:freeze_coco}
	\end{minipage}\hfill
	\vspace{-7mm}
\end{table}

%% file: figure/qualitative.tex
\begin{figure}
  \centering
  \includegraphics[width=1.0\linewidth]{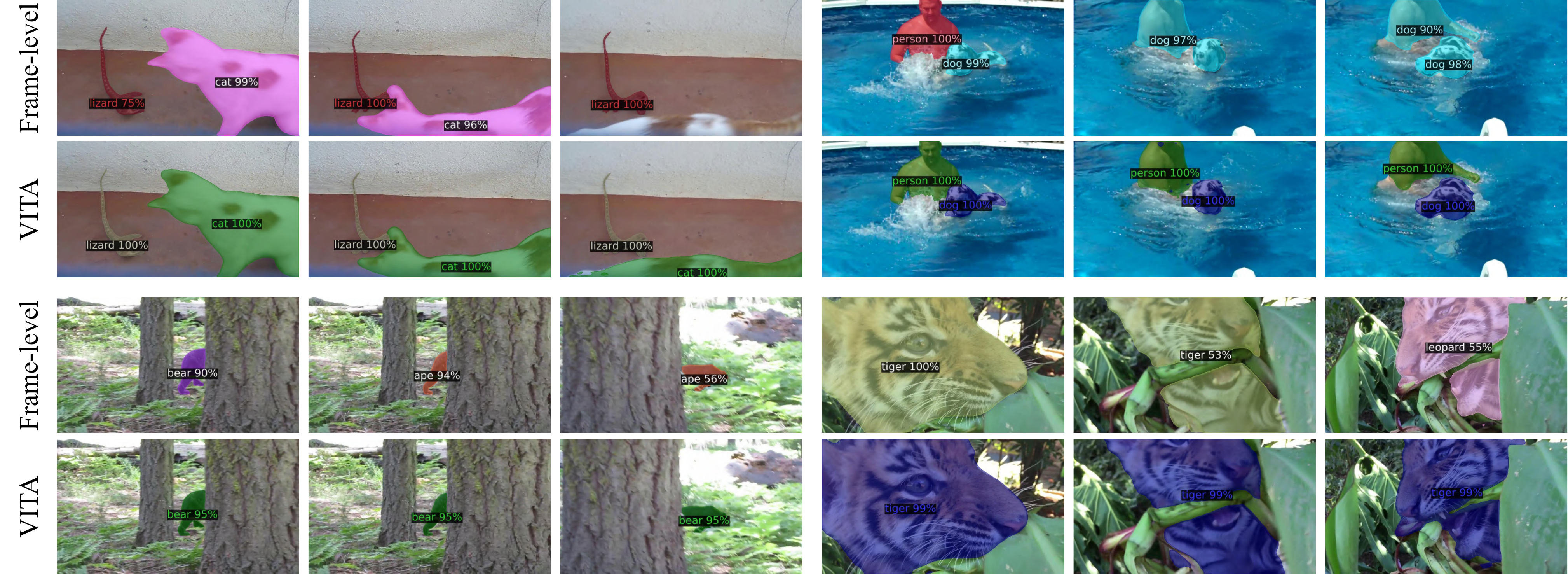}
  %\vspace{-5mm}
  \caption{
  %TODO: CNN->backbone; N_f and N_v; hat(f) and f; Mask feature M;
  Visualization of predictions from the frame-level detector and VITA. Instances with the same identity are displayed in the same color.
  }
  \label{fig:qualitative}
  %\vspace{-5mm}
\end{figure}

%% file: sections/5_conclusion.tex
\section{Limitations}
\label{sec:limitations}
VITA has achieved high performance in the complete-offline paradigm while dramatically improving the number of input frames that can be processed at once.
However, there are two major limitations for the ultimate long video understanding.
First, the current architecture still has limitations in processing an infinite number of frames.
In addition, since object tokens do not explicitly utilize temporal information, they may have difficulties in identifying complex behaviors that span over very long sequences.
We believe that devising explicit designs to address these issues will be a promising future direction.

\section{Conclusion}
\label{sec:conclusion}
In this paper, we proposed VITA for offline Video Instance Segmentation.
VITA is a simple model built on top of the off-the-shelf image instance segmentation model~\cite{Mask2Former}.
Unlike existing offline methods, VITA directly leverages object queries decoded by independent frame-level detectors.
We demonstrated that deploying object-oriented information is not only effective in improving performance, but also has robust practicality for processing long and high-resolution videos - setting state-of-the-art on popular VIS benchmarks, \emph{e.g.}, YouTubeVIS-2019 \& 2021 and OVIS.
Moreover, since VITA is designed to absorb spatial knowledge purely from image object detector, it shows fast convergence and demonstrates competitive performance even if trained on frozen detectors.
We hope that our method extends the scope of offline VIS research beyond benchmarks to real-world applications.
% For the ultimate long video understanding, we believe that devising an explicit design that takes temporal information will be a promising future research direction.

\section*{Acknowledgements}
\label{sec:acknowledgements}
% TODO: ADD continual learning
This work has partly supported by the National Research Foundation of Korea (NRF) grant funded by the Korea government (MSIT) (NRF2022R1A2C2004509) and by Institute of Information communications Technology Planning Evaluation (IITP) grant funded by the Korea government (MSIT) (No. 2022-0-00113, Developing a Sustainable Collaborative Multi-modal Lifelong Learning Framework), and Artificial Intelligence Graduate School Program under Grant 2020-0-01361.

\section*{Broader Impact}
\label{sec:broader impact}
VITA is designed for the VIS task and focuses on processing long and high-resolution videos in an end-to-end manner while achieving the state-of-the-art performance.
We hope that VITA can have a positive impact on many industrial areas such as video editing applications.
We would like to note that research on VIS must be aware of potential misuse that violates personal privacy.
\paragraph{Licenses of COCO~\cite{COCO}, YouTube-VIS~\cite{MaskTrackRCNN}, OVIS~\cite{OVIS-Dataset}, and  \texttt{detectron2}~\cite{Detectron2}:} Attribution 4.0 International, CC BY 4.0, CC BY-NC-SA 4.0, and Apache-2.0, respectively. 
%%% LICENSES %%%
% COCO - https://cocodataset.org/#termsofuse
% OVIS - https://competitions.codalab.org/competitions/32377#learn_the_details-terms_and_conditions
% YouTubeVIS - https://youtube-vos.org/dataset/term/
% Detectron2 - https://github.com/facebookresearch/detectron2

%% file: sections/6_supplementary.tex
% Supplementary에 pseudo코드같은거로 왜 우리거는 long&high-res vid 되고 mask2former는 안되는지 비교하면 좋을 것 같다.
% Ablation Study 에서 Maximum number of frames 구할 때 다른 방법 어떻게 수정했는지에 대한 디테일
% Frozen Detector: COCO 랑 안겹치는 클래스에 대한 qualitative result 필요
% Mask2Former decoder 9 layer중에서 3개만 썻다
% 4 A100 with AMP training (8 for Swin)
% Pseudo Video 만드는 과정
% YTVIS 2019에서는 다섯프레임, OVIS, YTVIS 2021에서는 여섯프레임

% 서플 기간에 챌린지 참여도 해야함...

% source code: 다는 아니더라도 메인 부분만이라도.
% conventional 한 방법들 대비 한번에 in/out 되는 inference code
% input 그냥 zero로 해서 프레임 넣어볼 수 있도록.

% youtube2019 all frames inference 해서 visualization
% (다른 메소드의 5프레임짜리 인퍼런스 결과)

% ablation heuristic clip association 에서 우리 방법 내에서 heuristic 들어갔을 때 일어나는 시나리오 보여주는 것
\newpage
\appendix
\section*{Appendix}
%\subsection*{VITA: Video Instance Segmentation via Object Token Association}
% Our supplementary materials consist of (1) a document, (2) a visualization video file, (3) and source codes.
In this Appendix, we first provide more training details of VITA (\cref{supp:training}). 
In addition, the inference procedure is explained in \cref{supp:inference}.
% Then, we provide brief information about our source code in \cref{supp:code}.
% Finally, in \cref{supp:visualization}, we explain the attached video file.

\section{Training Details}
\label{supp:training}
\subsection{Implementation}
We use 4 NVIDIA A100 GPUs with 40GB of memory (8 A100 GPUs when using a Swin~\cite{Swin} backbone), and activate Automatic Mixed Precision (AMP) provided by PyTorch.
% For training, we use 4 Nvidia A100 gpus with 40GB GPU memory (8 A100 when using Swin~\cite{Swin}), and enable Automatic Mixed Precision (AMP) provided by PyTorch.
Our training pipline is two-stage.
We first pretrain the model for image instance segmentation on COCO~\cite{COCO} \texttt{train} set using the batch size of 16 and by setting the number of input frames to $T=1$.
%We use AdamW optimizer with initial learning rate of $10^{-5}$ for backbone and $10^{-4}$ for the rest of the modules.
Then, we finetune the pretrained model on the VIS \texttt{train} sets (YouTube-VIS 2019~\cite{MaskTrackRCNN}, YouTube-VIS 2021, and  OVIS~\cite{OVIS-Dataset}) with pseudo-videos augmented from COCO images (see \cref{supp:pseudo-video}).
We use the batch size of 8 and set each input clip to be length of $T=6$.
Considering the difficulty and varying number of training videos included in each dataset, we set up different training iterations for each VIS dataset - 130k, 160k, 110k with decay of learning rates at 75k, 100k, 50k for YouTube-VIS 2019, 2021, and OVIS, respectively.
And both Object Encoder and Object Decoder in VITA follow the standard Transformer encoder and decoder architectures suggested in DETR~\cite{DETR}.
However, we just switch the order of self- and cross-attention in Object Decoder to make video queries learnable, and eliminate dropouts to make computation more efficient, as discussed in Mask2Former~\cite{Mask2Former}.

\subsubsection{Pseudo-video generation}
\label{supp:pseudo-video}
During training, we follow SeqFormer~\cite{SeqFormer} to generate pseudo-videos from a single image.
Given a single image, we first resize the short side of an image to one of the 400, 500 and 600 pixels while maintaining its ratio.
Then, the image is randomly cropped $T$ times to a size in the range [384, 600] to create a pseudo-video of length $T$.
% To begin with, an image is resized with its ratio kept and a shorter edge size to be randomly chosen from 400, 500, and 600 pixels.
%Then, a crop of random size within the range [384, 600] is applied to the image.
Finally, the cropped images are resized to a shorter edge to be randomly chosen from [288, 512] pixels with a step of 32 pixels.

\subsection{Loss function}
\label{supp:loss}
The final loss function of our frame-level detector~\cite{Mask2Former}, denoted by $\mathcal{L}_{f}$ in the main paper, is largely composed of two terms: mask-related loss and categorical loss.
The mask-related loss is again consists of $\mathcal{L}_{ce}^{f}$ and $\mathcal{L}_{dice}^{f}$, each representing a binary cross-entropy loss and a dice loss, respectively.
Then, the final loss $\mathcal{L}_{f}$ is a combination of a categorical loss (the cross entropy) and the mask-related loss $\mathcal{L}_{f} = \lambda_{cls}\mathcal{L}_{cls}^{f} + \lambda_{ce}\mathcal{L}_{ce}^{f} + \lambda_{dice}\mathcal{L}_{dice}^{f}$ and we set $\lambda_{cls} = 2, \lambda_{ce} = 2$, and $\lambda_{dice} = 5$, respectively.

For the $\mathcal{L}_{v}$ calculated from video-level results generated by VITA, we employ the same hyper-parameters as frame-level losses: $\mathcal{L}_{v} = \lambda_{cls}\mathcal{L}_{cls}^{v} + \lambda_{ce}\mathcal{L}_{ce}^{v} + \lambda_{dice}\mathcal{L}_{dice}^{v}$. Note that, for $\mathcal{L}_{ce}^{v}$ and $\mathcal{L}_{dice}^{v}$, we extend the functions of $\mathcal{L}_{ce}^{f}$ and $\mathcal{L}_{dice}^{f}$ to the temporal axis, just as IFC~\cite{IFC} did.

\subsection{Building VITA on Mask2Former}
\label{supp:vita}
Mask2Former uses 9 decoder layers where output frame queries from each layer can be used as an input for VITA.
However, using the outputs from all 9 layers during training leads to the lack of GPU memory.
Therefore, we use the outputs from the last 3 layers for training VITA.

\section{Inference procedure}
\label{supp:inference}
In~\cref{tab:max_frame} in the main paper, we measured the maximum number of frames that each model can infer at once.
To further specify the process of measuring the numbers, we provide simplified PyTorch-style inference pseudo-codes of both VITA and Mask2Former-VIS in~\cref{alg:vita} and~\cref{alg:mask2former} respectively.
For fair comparison, we modified the inference procedure of previous methods to collect backbone features of each frame sequentially. 
The strategy prevents the methods from a memory explosion until entering each VIS prediction module.
The most noticeable difference is that VITA collects only \texttt{frame_queries} and \texttt{mask_features} of each frame from our frame-level detector~\cite{Mask2Former} denoted by the function \texttt{mask2former()} (line 2-12 in~\cref{alg:vita}).
Then, the \texttt{frame_queries} for the entire video become the input of Object Encoder (line 19 in~\cref{alg:vita}).
On the other hand, previous Transformer-based offline VIS models (\textit{e.g.}, Mask2Former-VIS), first aggregate the backbone features of entire video and takes it as inputs for the VIS model, the function \texttt{mask2former_vis()} (line 3-20 in~\cref{alg:mask2former}).
After that, both of methods generate their video-level predictions by using their \texttt{vq} (video queries) and \texttt{mask_features}.

% \section{Code}
% \label{supp:code}
% We include the source code to the zip file.
% Further implementation details of the VITA module can be found in \texttt{vita.py}, and the training and inference functions are included in \texttt{meta_arch.py}.
% Upon acceptance, all codes and environmental guidelines for training \& inference will be made publicly available, including trained checkpoints.

\input{algorithm/inference}
% \section{Visualization videos}
% \label{supp:visualization}
% \subsection{Frozen detector}
% In \cref{tab:freeze_coco} of main paper, our VIDI with frozen detector showed competitive quantitative results on YouTube-VIS 2019 benchmark.
% We visualize results for videos that contain instances in semantic categories that are only included in YouTube-VIS 2019 and do not overlap with COCO.
% Interestingly, VIDI effectively predicts non-overlapping categories under various scenarios (see 00:05 - 00:38 in the attached video file).

% \subsection{YouTube-VIS 2019 all frames}
% The videos in YouTube-VIS 2019 \texttt{valid} dataset consist of sampling 1 out of 5 frames of the original video.
% Therefore, the original maximum number of frames of \texttt{valid} videos is 180, which cannot be processed at once by existing offline VIS methods using normal GPUs (see \cref{tab:max_frame} in the main paper).
% We provide qualitative results of VITA on both \texttt{valid} dataset and the original \emph{all frames} videos (see 00:39 - 01:26 in the attached video file).
% Note that VITA infers the \emph{all frames} videos without any hand-crafted post-processing algorithm.

%% file: algorithm/inference.tex
\begin{table}[H]
	\centering
	\begin{minipage}[t]{0.49\linewidth}
    \caption{
    PyTorch-style inference pseudo-code of VITA.
    }
    \label{alg:vita}
    \begin{python}
def vita(video):
    frame_queries = []
    mask_features = []

    for frame in video:
        feats = backbone(frame)
        fq, mf = mask2former(
            feats
        )

        frame_queries.append(fq)
        mask_features.append(mf)

    """
    VITA only aggregates
    frame queries for its
    remaining computations.
    """
    fq = object_encoder(
        frame_queries
    )
    vq = object_decoder(fq)

    w = mask_head(vq)
    pred_mask = []
    for mf in mask_features:
        # w.shape: (Nv x C)
        # mf.shape: (C x H x W)
        _mask = w @ mf

        pred_mask.append(_mask)

    # Nv x (K+1)
    pred_cls = cls_head(vq)

    # Nv x T x H x W
    pred_mask = torch.stack(
        pred_mask, dim=1
    )

    return pred_cls, pred_mask
    \end{python}
    %\vspace{-9mm}
	\end{minipage}
	\hfill
	\begin{minipage}[t]{0.49\linewidth}
    \caption{
    PyTorch-style inference pseudo-code of Mask2Former-VIS~\cite{Mask2Former-VIS}.
    }
    \label{alg:mask2former}
    \begin{python}
def previous_methods(video):

    frame_features = []

    for frame in video:
        feats = backbone(frame)
        frame_features.append(
            feats
        )

    """
    Previous approaches receive
    either multi or single scale
    feature map at once for their
    encoder/decoder layers.
    """
    vq, mask_features =\
        mask2former_vis(
            frame_features
        )

    w = mask_head(vq)
    pred_mask = []
    for mf in mask_features:
        # w.shape: (Nv x C)
        # mf.shape: (C x H x W)
        _mask = w @ mf

        pred_mask.append(_mask)

    # Nv x (K+1)
    pred_cls = cls_head(vq)

    # Nv x T x H x W
    pred_mask = torch.stack(
        pred_mask, dim=1
    )

    return pred_cls, pred_mask
    \end{python}
    %\vspace{-9mm}
	\end{minipage}
\end{table}